\documentclass{article}

\PassOptionsToPackage{numbers}{natbib}
\usepackage[final]{neurips_distshift_2022}

\usepackage[utf8]{inputenc} 
\usepackage[T1]{fontenc}    
\usepackage{hyperref}       
\usepackage{url}            
\usepackage{booktabs}       
\usepackage{amsfonts}       
\usepackage{amsmath}
\usepackage{amssymb}
\usepackage{mathtools}
\usepackage{amssymb}
\usepackage{bbm}
\usepackage{dsfont}
\usepackage{nicefrac}       
\usepackage{microtype}      
\usepackage[dvipsnames]{xcolor}
\usepackage{tikz}
\usepackage{floatrow}
\usepackage[font=small]{caption}
\usepackage[bottom]{footmisc}
\usepackage{algorithm}
\usepackage{algpseudocode}
\usepackage[shortlabels]{enumitem}

\newtheorem{Assumption}{Assumption}

\def\x{{\mathbf{x}}}
\def\w{{\mathbf{w}}}

\def\bphi{{\boldsymbol{\phi}}}
\def\R{{\mathbb{R}}}

\usepackage{amsmath}

\DeclareMathOperator*{\argmin}{argmin}

\newcommand{\youngsuk}[1]{}
\newcommand{\michael}[1]{}

\newcommand{\dataset}[1]{\texttt{#1}}

\newcommand{\covid}{\dataset{Covid-deaths}}
\newcommand{\electricity}{\dataset{Electricity}}
\newcommand{\traffic}{\dataset{Traffic}}
\newcommand{\solar}{\dataset{Solar}}
\newcommand{\taxi}{\dataset{Taxi}}

\newcommand{\model}[1]{\texttt{#1}}
\newcommand{\adaptiveDeepAR}{\model{Ada-DeepAR}}
\newcommand{\adaptiveDeepARuniform}{\model{Ada-DeepAR}}
\newcommand{\deepar}{\model{DeepAR}}
\newcommand{\revin}{\model{RevIn}}
\newcommand{\tft}{\model{TFT}}

\title{Adaptive Sampling for Probabilistic Forecasting under Distribution Shift}

\author{%
  Luca Masserano\thanks{Work done while at AWS AI Labs} \\
  Department of Statistics\\
  Carnegie Mellon University\\
  \texttt{lmassera@andrew.cmu.edu} \\
  \And
  Syama Sundar Rangapuram\thanks{Correspondence to: rangapur@amazon.com} \\
  AWS AI Labs\\
  \texttt{rangapur@amazon.com} \\
  \And
  Shubham Kapoor \\
  AWS AI Labs\\
  \texttt{kapooshu@amazon.com} \\
  \And
  Rajbir Singh Nirwan$^*$\\
  D2S Inc. \\
  \texttt{rajbir.nirwan@gmail.com} \\
  \And
  Youngsuk Park\\
  AWS AI Labs\\
  \texttt{pyoungsu@amazon.com} \\
  \And
  Michael Bohlke-Schneider\\
  AWS AI Labs\\
  \texttt{bohlkem@amazon.com} \\
}

\begin{document}
\maketitle

\begin{abstract}

The world is not static: This causes real-world time series to change over time through external, and potentially disruptive, events such as macroeconomic cycles or the COVID-19 pandemic. We present an adaptive sampling strategy that selects the part of the time series history that is relevant for forecasting. We achieve this by learning a discrete distribution over relevant time steps by Bayesian optimization. We instantiate this idea with a two-step method that is pre-trained with uniform sampling and then training a lightweight adaptive architecture with adaptive sampling. We show with synthetic and real-world experiments that this method adapts to distribution shift and significantly reduces the forecasting error of the base model for three out of five datasets.

\end{abstract}

\section{Introduction}
\label{sec: intro}

Time series forecasting uses historical data to forecast the evolution of a time series to assist decision making in several domains, for example retail~\cite{mukherjee2018armdn, Croston1972, taghizadeh2017utilizing}, electric load planning~\cite{Saxena2019, HONG2019}, cloud computing management~\cite{park2019linear,park2020structured}, or labor planning~\citep{bohlkem2020}. Most time series forecasting algorithms make the assumption that the data distribution does not change over time. However, this is violated in practice where real-world time series data is affected by drifts such as evolving customer demand or disruptive events (like the COVID-19 pandemic). These distribution shifts will inevitably be part of the time series history and therefore forecasting algorithms need to be able to account for them. 

Distribution shifts can occur with different (possibly reoccurring) patterns, but likely result in higher test error because the train and test distribution differs. Thus, only a subset of the time series history might come from the same distribution as the test set. Driven by this observation, we propose an \textit{adaptive sampling mechanism} that explicitly feeds the model with inputs from the history that are only relevant for forecasting in the current (most recent) distribution. Instead of sampling uniformly over the history, our method learns a discrete distribution over past time steps and explicitly allows continuous adaptation, so that the model can react and adapt to distribution shifts. 

Our contribution is an adaptive sampling mechanism that learns the sampling distribution over time steps via gradient-free Bayesian optimization. We propose a two-step framework where the forecasting model is pre-trained with uniform sampling and then a lightweight architecture is trained with adaptive sampling. We evaluate the properties and performance of this approach with experiments on synthetic and real-world data. This paper is organized as follows: Section~\ref{sec: background} introduces the reader to the background of this paper and the types of data distribution shifts for forecasting that we address. and Section~\ref{sec: model} introduces our method and we present experiments in Section~\ref{sec: experiments}. Section~\ref{sec: conclusion} concludes the paper. Note that the discussion of the related work can be found in Section~\ref{sec: related_work} in the Appendix.

\section{Background}
\label{sec: background}

\begin{figure}[t!]
    \centering
    \includegraphics[width=0.9 \textwidth]{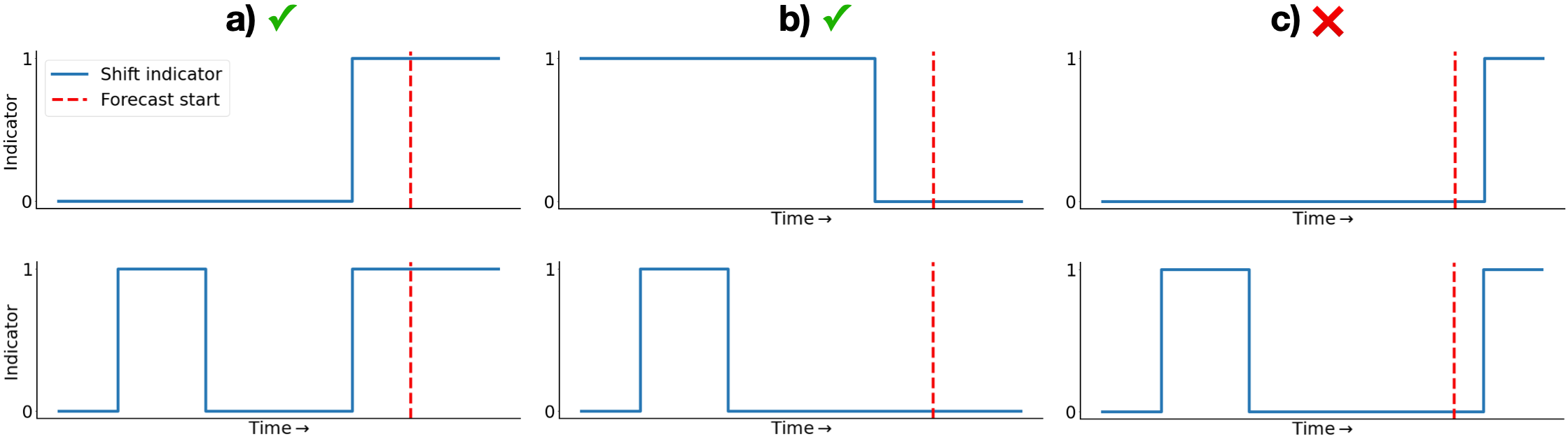}
    \caption{\small
    \textbf{Data distribution shift that our model can address.} Each frame shows an indicator function plotted over time with respect to an assumed “standard” distribution (a value of $1$ indicates a distribution shift). The red dashed line separates the past (i.e., the training set) from the future (i.e. test set). \textbf{a)} New or reoccurring shift: the distribution over time steps should focus on the shifted regions. \textbf{b)} This case is complementary to the previous one: here the distribution over time steps should exclude the shifted regions. \textbf{c)} The shift occurs in the forecasting window but not in the most recent region before the forecast start date: our model needs to wait until part of the shift is in the training set in order to be able to adapt to it.
    }
    \label{fig:shift_cases}
\end{figure}

\paragraph{Notation} We denote the value of time series $i$ at time $t$ by $z_{i, t}$, where $z_{i, t} \in \R$, and corresponding time-varying covariates as $\x_{i, t} \in \R^d$.
The past of each time series (i.e., training set) is denoted as $z_{i, 1:T} \coloneqq (z_{i, 1}, \dots, z_{i, T})$, and the future (i.e., forecasting window or test set) as $z_{i, T+1: T+\tau} \coloneqq (z_{i, T+1}, \dots, z_{i, T+\tau})$. The number of time series in the dataset is denoted by $N$. 

Our main assumption in this work is that we are able to observe at least some of the forthcoming shift in order to be able to react and adapt to it: 

\label{sec: assumptions}
\begin{Assumption}
    Let $c$ be a constant number of (past) time steps set a priori. We assume that there is no (marginal) distribution shift between the most recent training window of $c$ steps and the forecasting (test) window. More precisely, let $z_{i, t}$ be the given time series at time $t$ and let $\bar{z}_{i, t}$ be the time series at time $t$ obtained after removing trend, seasonalities and other effects of available features, then
    \begin{equation}
        \label{eq: shift_assumption}
        p(\bar{z}_{i, T-c}) = \cdots = p(\bar{z}_{i, T}) = p(\bar{z}_{i, T+1}) = \cdots = p(\bar{z}_{i, T+\tau}), \quad \forall i = 1, \dots, N.
    \end{equation}
\end{Assumption}

This assumption implies that the data with $t \in [T-c, T]$ is the closest in time to the forecasting window, therefore it is reasonable to regard it as the most similar to the future. In practice, we take $c$ to be equal to the sum of context and prediction lengths, so that we can use $\{z_{i, T-c:T}\}_{i=1}^N$ as a validation set to learn and adapt to any distribution shift. In general, we consider the following two scenarios illustrated in Figure~\ref{fig:shift_cases}): \textbf{(a)} A new or reoccurring shift in the past and no shift between the recent training and test window; and \textbf{(b)} a shift that occurred in the past but reverted back to the "standard" regime.  Note that we do not address the case where the distribution shift occurs in the test window only, which would be more of a fundamental generalization problem (case \textbf{(c)} in Figure~\ref{fig:shift_cases}). See Section~\ref{sec: shift_types} in the Appendix for a detailed definition of the cases that we consider.

\section{Our Method}
\label{sec: model}
Training of deep learning models in time series settings is usually done by sampling fixed-size windows \textit{uniformly at random} over the available time series history (i.e., the past). Since distribution shifts happen with different (possibly reoccurring) patterns over time, uniform sampling over the training set would likely feed the algorithm with irrelevant or even misleading data to forecast in the current distribution regime. 
Therefore we propose to adaptively generate these training windows depending on the regime of the validation set (and by Assumption~\ref{eq: shift_assumption}, the test set). The key element of our approach is automatically learning which regions to focus on to accurately forecast the test set.

To achieve this, we introduce a discrete distribution $p_\bphi(\cdot)$ over time steps $1, \dots, T$ that determines the probability of sampling a training window starting at a time point $t$. In practice, $p_\bphi(\cdot)$ can be any discrete distribution that is well defined over integers, such as Geometric, Negative Binomial, Poisson, or mixtures of these distributions, parameterized by $\bphi$. This distribution is responsible for sampling time windows that are used to train our forecasting model (see Figure~\ref{fig: ada_sampling} in the Appendix for an illustration). 
We learn $\bphi$ by minimizing validation loss of the model that is trained using examples generated according to $p_\bphi(\cdot)$.
Ideally, the learned distribution parameters $\bphi^\star$ will avoid sampling distribution shifts in history that do not correspond to the current regime. This distribution replaces uniform sampling over the history, which cannot distinguish between distribution shifts and normal regions. However, our model \emph{should} fall back to uniform sampling if this results in lowest validation error. This suggests that no distribution shift is present in the validation window and consequently in the forecast window (by Assumption~\ref{eq: shift_assumption}). 

\paragraph{Adaptive training of a lightweight architecture} 
\label{sec: ada_training}
In practice, we would expect to frequently re-estimate $\bphi^\star$ to adapt to the current distribution. To address this, we propose a two-step procedure with a lightweight adaptive architecture that is fast to train and can be frequently re-trained. In this sense, this approach is explicitly considering a notion of time-varying parameters.
\textbf{i)} The model is pre-trained using uniform sampling over the whole training history. Intuitively, this step aims at learning common features that are useful everywhere in time. \textbf{ii)} After freezing the weights of the pre-trained model, an adaptive lightweight architecture is attached and trained using the adaptive sampling mechanism, where the training windows are generated according to $p_{\bphi^*}(\cdot)$. The goal of this step is to learn time-specific features that are relevant to forecast in the current distribution. The lightweight architecture can be a short sequence of fully connected layers, or it can even be a subset of the pre-trained model. 

 \begin{figure}[t!]
     \floatbox[{\capbeside\thisfloatsetup{capbesideposition={right,top}}}]{figure}[\FBwidth]
     {\caption{\textbf{Schematic description of the two-step procedure.} \textbf{i)} Pre-train the full model with uniform sampling; \textbf{ii.a)} use Bayesian optimization to find the optimal $\bphi^\star$; \textbf{ii.b)} use $\bphi^\star$ to train the lightweight architecture with adaptive sampling.
     }
     \label{fig: ada_training}}
     {\includegraphics[width=0.68\textwidth]{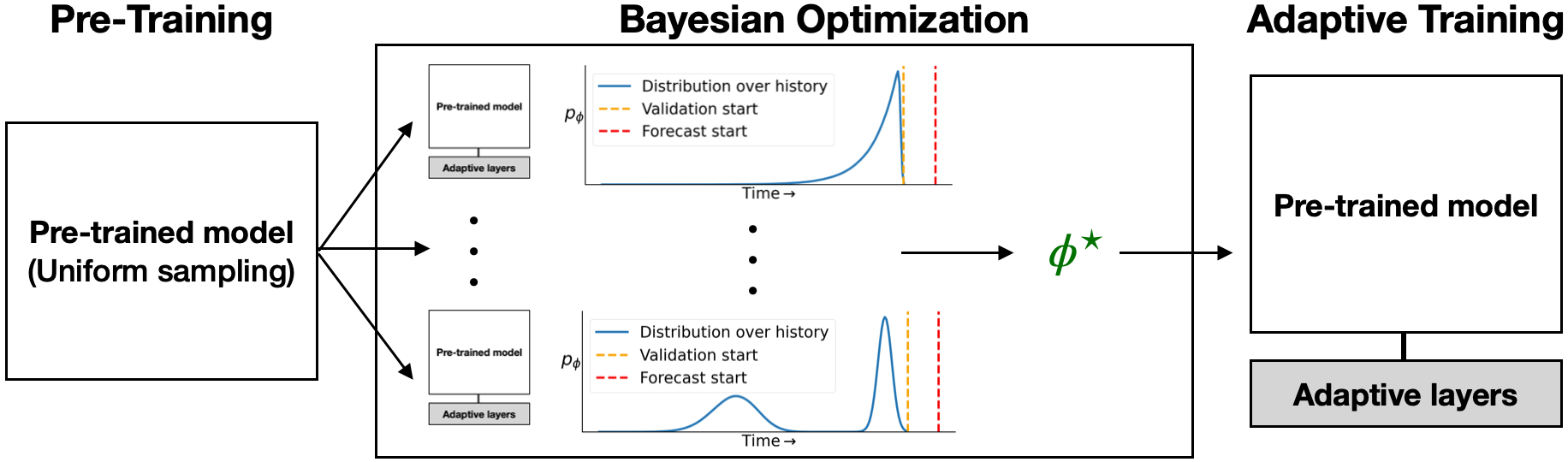}}
 \end{figure}

\paragraph{Learn $\bphi$ via Bayesian Optimization} 
We now describe how we estimate the parameters $\phi$ of the distribution $p_\bphi(\cdot)$. 
Let us denote the parameters of the pre-trained model as $\w_{\text{pre}}$ and those of the lightweight architecture as $\w_{\text{ada}}$. 
Moreover, let $f_\w(\{z_{i, 1:T}\})$ denote the forecasts $\{\hat{z}_{i, T+1:T+\tau}\}$ of the model whose weights are given by $\w$.
Given input time series $\{z_{i, t}\}, i= 1, \ldots, N$ and $t = 1, \ldots, T$, we set aside the last $\tau$ observations $\{z_{i, T-\tau + 1 : T}\}, \forall i$, for estimating $\phi$.
We learn $\phi$ by minimizing loss on this validation channel of the model trained according to $p_\bphi(\cdot)$. 
More precisely, we have
\[
    \bphi^* = \argmin_\bphi \ell(\bphi),
\]
where $\ell(\bphi)$ is the validation loss of the model that is trained using adaptive sampling $p_\bphi(\cdot)$,
\[
    \ell(\bphi) = \texttt{loss}\Big(f_{\w_{\text{pre}} \cup \w^*_{\text{ada}}(\bphi)}\big(\{z_{i, 1:T-\tau}\} \big), \{z_{i, T-\tau+1:T}\}\Big).
\]
Here $\w^*_{\text{ada}}(\bphi)$ are the weights obtained after fine-tuning the pre-trained model on the input data $\{z_{i, 1:T - \tau}\}$ where the training examples are generated according to the distribution $p_\bphi(\cdot)$; note the dependence of $\w^*_{\text{ada}}$ on $\bphi$.
A crucial observation is that one can only evaluate $\ell(\bphi)$ and cannot obtain its gradients with respect to $\bphi$; hence we cannot train $(\w_{\text{ada}}, \bphi)$ jointly end-to-end. 
We propose to treat $\bphi$ as hyper-parameters and use Bayesian optimization \cite{frazier2018tutorial} to find $\bphi^*$. Assumption~\ref{eq: shift_assumption} plays a key role in this step because the most recent region in the training set is used as validation channel to choose $\bphi^\star$ during Bayesian optimization. The overall approach is depicted in Figure~\ref{fig: ada_training} and the implementation details with \texttt{DeepAR}~\cite{salinas2020deepar} as the base model are given in Algorithm~\ref{algo:adaptiveDeepAR}. 
In our experiments we assume $\w_\text{ada} \subset \w_\text{pre}$ and hence the algorithm is presented for this case.

\section{Experiments}
\begin{figure}[t!]
    \floatbox[{\capbeside\thisfloatsetup{capbesideposition={right,top}}}]{figure}[\FBwidth]
    {\caption{\textbf{Experiments with synthetic data.} \textbf{a}) Synthetic data with injected shift (top) and noise (bottom). \textbf{b1)} (top case in \textbf{a}): Adaptive sampling focuses on the shifted region in the past (red and green bars denote the shifted region) that is relevant for the current regime.~\textbf{b2} (bottom case in \textbf{a}): Adaptive sampling avoids the injected noise shifts.
    \vspace{-5pt}
    }
    \label{fig: results}}
    {\includegraphics[width=0.7\textwidth]{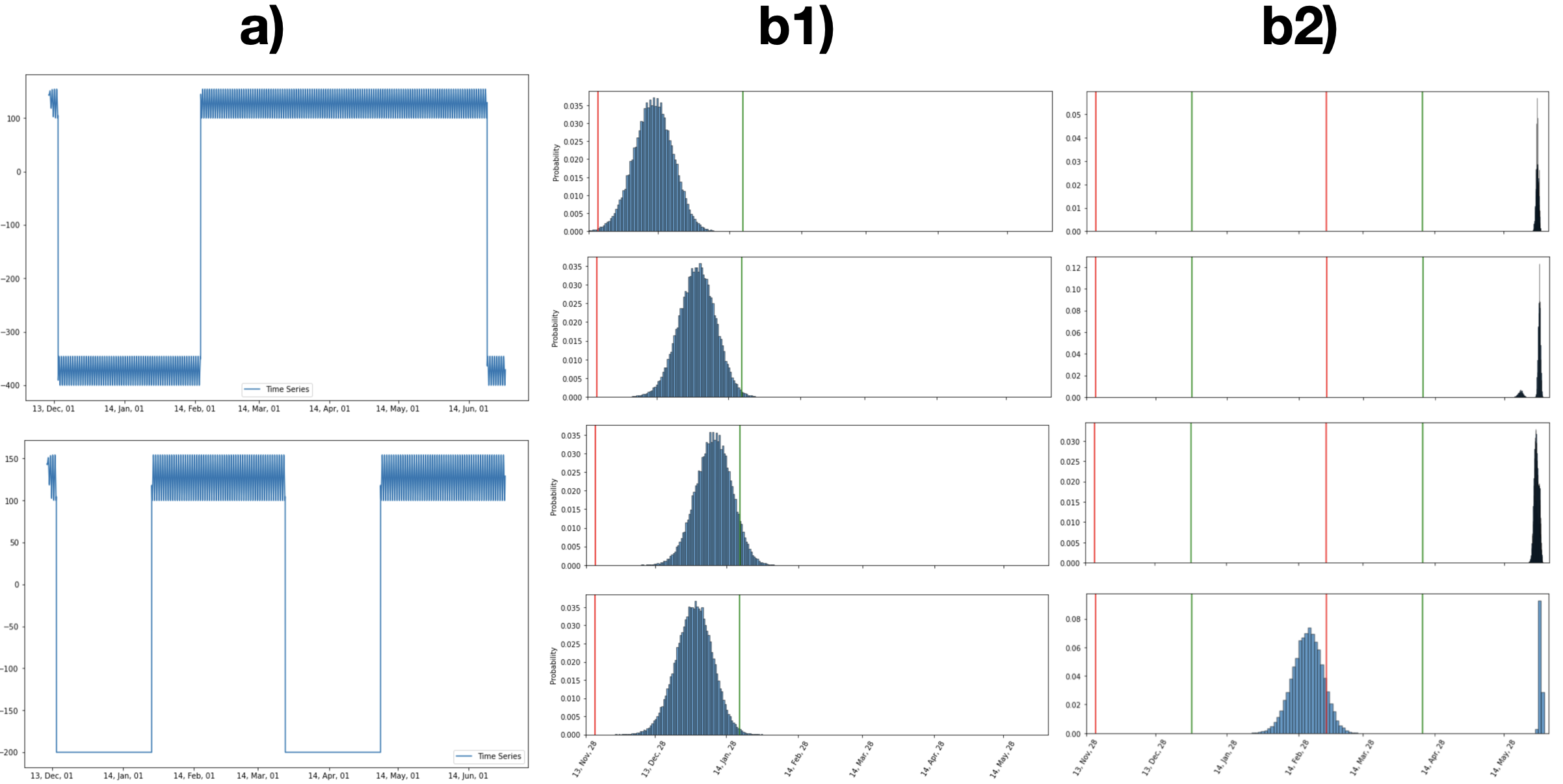}}
\end{figure}

\begin{algorithm}
    \caption{Adaptive DeepAR}
    \label{algo:adaptiveDeepAR}
    \begin{algorithmic}[1] 
    \small
        \Procedure{\adaptiveDeepAR}{$\{z_{i, t}\},\ i \in [1 \ ..\ N],\ t  \in [1 \ .. \ T]$}
            \State $Z_{\text{train}} = \{z_{i, t}\},\forall i,\ \forall t$
            \State $\w_{\text{pre}} \gets \deepar(Z_{\text{train}})$
            \Comment{Standard training with uniform train sampler}
            \newline 
            \State $Z_\text{ada\_train} = \{z_{i, t}\},\ \forall i,\ t \in [1 \ .. \ T - \tau]$
            \State $Z_\text{validation\_labels} = \{z_{i, t}\},\ \forall i,\ t \in [T - \tau + 1\ .. \ T]$
            \State $\bphi = \bphi_0$  
            \Comment{Random initialization}
            \State $\w = \w_{\text{pre}}$
            \State Split $\w$ into $\w_{\text{ada}}$ and $\w_{\text{frozen}}$
            \While{\text{CONVERGENCE}} 
                \Comment{Solve $\min_\bphi \ell(\bphi)$}
                \State $\w_{\text{ada}} \gets \deepar(Z_\text{ada\_train}, \text{init}=\w, \text{sampler}= p_\bphi(\cdot), \text{fixed}=\w_{\text{frozen}})$
                \Comment{Train with adaptive train sampler by fixing weights $\w_{\text{frozen}}$}
                \State $\w =\w_{\text{ada}} \cup  \w_{\text{frozen}}$
                \State $\hat{Z}_\text{validation\_preds} = \text{forecast}(Z_\text{ada\_train}, \text{model}=\w)$ 
                \State $\ell(\bphi) = $ loss($\hat{Z}_\text{validation\_preds}, Z_\text{validation\_labels}$)
                \State update $\bphi$ based on the new evaluation $\ell(\bphi)$
                \Comment{Bayesian Optimization}
            \EndWhile\label{euclidendwhile}
            
            \Comment{$p_\bphi(\cdot)$ likely has more mass on indices $t$ that help predict $Z_{\text{validation\_labels}}$ accurately}
            \State \textbf{return} $\w$\Comment{Model is ready to predict $\{z_{i, t}\},\ \forall i,\ t \in [T+1\  ..\ T +\tau]$}
        \EndProcedure
    \end{algorithmic}
\end{algorithm}

The instantiation of our model in this manuscript is called \adaptiveDeepAR {} and uses \texttt{DeepAR}~\cite{salinas2020deepar} as a base model with a two-layer LSTM (other base models could be used). The adaptive part of our model is the second LSTM layer. Additionally, we perform Bayesian optimization using MOBSTER, which combines asynchronous successive halving and asynchronous Hyperband with gaussian-process based Bayesian optimization~\citep{klein2020mobster}. We use the implementation available in \texttt{Syne Tune}~\cite{salinas2022syne}. For any time series $i$ with training history $T$ and forecast horizon~$\tau$, we use $z_{i, T-\tau:T}$ as a validation channel for Bayesian optimization. We consider a geometric and a mixture of two negative binomial distributions as our sampling distributions $p_\bphi(\cdot)$. Training details are given in Section~\ref{sec: training_details}.

\paragraph{Experiments on synthetic data:}
To demonstrate the applicability of our method, we first evaluate it on two synthetic experiments that mimic cases \textbf{(a)} and \textbf{(b)} of Figure~\ref{fig:shift_cases}. We generated an artificial sinusoidal dataset and then injected shifts or noise in different regions. \textbf{Focus on shifted regions distant in time:} In this example we injected a shift in a distant portion of the training dataset and also in the validation and test sets (see the top of panel \textbf{a)} in Figure~\ref{fig: results}).  We show that the sampling distribution over time steps can focus on specific regions that are relevant to the current domain, regardless of when they occurred in the available history (see \textbf{b1)} in Figure~\ref{fig: results}). \textbf{Exclude noisy regions:} In this example, we injected noise (a negative constant) in two different regions of the training dataset. Our adaptive sampling mechanism avoids these noisy regions from the history (\textbf{b2)} in Figure~\ref{fig: results}). Interestingly, this sometimes corresponds to sampling only in the most recent window, while in other cases the distribution exploits its bimodality to focus on two distant windows.

\label{sec: experiments}
\begin{table*}[t]

\begin{small}
\begin{center}
\begin{tabular}{lll|lll}
\toprule
 &            \adaptiveDeepAR &                \adaptiveDeepAR-\texttt{uniform} & \revin        &  \deepar & \tft\\
Dataset               &   \texttt{(ours)}                        &       \texttt{(ablation)}                    &                     &  &     \\
\midrule
\covid         &  $0.074^{*}\pm$\tiny{$0.013$} &  $0.088\pm$\tiny{$0.011$} &   $0.13\pm$\tiny{$0.006$} & $0.103\pm$\tiny{$0.053$} & $\textbf{0.053}\pm$\tiny{$0.061$}\\
\electricity         &  $\textbf{0.048}^{*\dagger}\pm$\tiny{$0.002$} &  $0.054^{\dagger}\pm$\tiny{$0.005$} & $0.063\pm$\tiny{$0.005$}   &  $0.062\pm$\tiny{$0.011$} & $0.082\pm$\tiny{$0.000$}\\
\traffic &  $0.172\pm$\tiny{$0.002$} &  $0.171\pm$\tiny{$0.004$} &  $0.172\pm$\tiny{$0.003$}  & $0.176\pm$\tiny{$0.004$}  & $0.306\pm$\tiny{$0.050$}\\
\solar                &   $\textbf{0.370}^{*\dagger}\pm$\tiny{$0.003$} &  $0.376^{\dagger}\pm$\tiny{$0.003$} &  $0.376\pm$\tiny{$0.004$}  & $0.388\pm$\tiny{$0.007$} & $0.562\pm$\tiny{$0.015$}\\
\taxi                  &  $0.286^{\dagger}\pm$\tiny{$0.003$} &  $\textbf{0.278}^{*\dagger}\pm$\tiny{$0.001$} & $0.288\pm$\tiny{$0.004$} & $0.279\pm$\tiny{$0.004$}   & $0.509\pm$\tiny{$0.001$}\\
\midrule
\texttt{Mean nCRPS}   & $0.190$ & $0.193$ & $0.202$ & $0.206$ & $0.302$ \\
\texttt{Mean Rank}    &  $1.8$        & $1.8$         & $3.4$        & $3.4$         & $4.2$ \\

\bottomrule

\end{tabular}

 \caption{The nCRPS loss (lower is better) for all datasets and models. We average over ten runs and report the mean and standard deviation. We took the results for TFT from TSBench~\cite{tsbench}, which only uses two runs. Boldface numbers indicate that the differences in error between the best and the second-best method are statistically significant (paired t-test p-value $<0.05$). * indicates a statistically significant result when comparing \adaptiveDeepAR{} and \adaptiveDeepARuniform{}-\texttt{uniform}. $\dagger$ indicates a statistically significant result when comparing \adaptiveDeepAR{} or \adaptiveDeepARuniform{}-\texttt{uniform} to its base model \deepar.
 \vspace{-5pt}
 }
 \label{tbl:results}
\end{center}
\end{small}
\end{table*}

\youngsuk{Recommend to visualize the predictions and see if there is notable and qualitative differences in the context of distribution shift beyond just the quantitative improvement in error metric}
\michael{Thank you for the comment - I doubt we will have time for it for the workshop submission}
\paragraph{Experiments on real-world datasets:} Finally, we show preliminary results on five real-world datasets (see Section~\ref{sec: datasets} for details). We compare the following models: our two-step training procedure with adaptive sampling (\adaptiveDeepAR), reversible instance normalization (\revin)~\cite{kim2021reversible}, and the attention-based Temporal Fusion Transformer \tft{}~\cite{LIM20211748}. 
We also consider an ablation variant of our method,  \adaptiveDeepAR-\texttt{uniform}, which uses the same architecture and two-step procedure as \adaptiveDeepAR{}, but performs adaptive training (i.e., fine-tuning of the adaptive weights $\w_{\text{ada}}$) using uniform sampling instead of $p_{\bphi}(\cdot)$.
This variant is included to highlight the effectiveness of learning the sampling distribution $p_{\bphi}$ by isolating the gains obtained by our model due to the fine-tuning step. 
For \adaptiveDeepAR{}, \adaptiveDeepAR-\texttt{uniform} and \revin{} we use \deepar{}\cite{salinas2020deepar} as the base model. 
 \adaptiveDeepAR-\texttt{uniform} is different from \deepar{} because of the fine-tuning step where adaptive weights are again trained one more time; this might help correcting the overfitting of the \deepar{} model  (and reduce test error) because of the lower capacity of adaptive weights and/or help escape local minima (and reduce training loss).
We evaluate these probabilistic methods using the continuous-ranked probability score (CRPS)~\cite{crps1, crps2}. Table~\ref{tbl:results} summarizes the results. Averaged over all datasets,~\adaptiveDeepAR{} improves over the base model~\deepar{} by 8.4\% (relative improvement). For three out of five datasets, the improvements over the base model are statistically significant (paired Student's $t$-test $p$-value $<0.05$, see Table~\ref{tbl:results}).~\adaptiveDeepAR{} selected the mixture of two negative binomial distributions for all datasets except \covid{}~(for which the geometric distribution was selected). \revin{} also improves over \deepar{}, but with a much smaller margin.~\tft{} performs best on \covid{}, but overall worse than the other methods on these datasets. 
Interestingly, \adaptiveDeepAR-\texttt{uniform} also improves the error by 6.7\% over its base model. In particular, \adaptiveDeepAR-\texttt{uniform} is the best model on the \traffic{} and \taxi{} datasets. Our hypothesis is that adaption to distribution shift is not necessary, and perhaps even counterproductive, for these datasets. This is also suggested by the performance of \revin{}, which performs similar to \adaptiveDeepAR{} on these datasets. Note that a uniform distribution could be seen as one of the possible distributions $p_\bphi(\cdot)$ that our adaptive architecture should consider and if no other distribution reduces the validation error, the uniform distribution \emph{should} be selected. The performance differences between \adaptiveDeepAR{} and \adaptiveDeepAR-\texttt{uniform} are statistically significant for all datasets except \traffic.

\section{Conclusion}
\label{sec: conclusion}
We introduced an adaptive sampling mechanism over the training history, whose parameters are learned using Bayesian optimization. Our experiments show that our method can improve the forecasting error over its base model (in this work, the improvements over the base model where statistically significant for three out of five datasets). While we find Bayesian optimization to be a powerful technique to find good adaptive sampling parameters, this approach is computationally expensive (because many parallel trials need to be evaluated) and could degrade as the number of parameters increases. Re-framing our approach as an input-selection layer that uses the re-parametrization trick for sampling could allow for cheaper end-to-end learning of the adaptive model parameters and $p_\bphi(\cdot)$.



\newpage
\bibliographystyle{plainnat}
\bibliography{references}
\newpage
\appendix
\section{Appendix}
\setcounter{figure}{0}
\renewcommand\thefigure{\thesection.\arabic{figure}} 
\setcounter{table}{0}
\renewcommand\thetable{\thesection.\arabic{table}} 

\subsection{Related Work}
\label{sec: related_work}

The problem of learning under distribution shift for time series has been extensively studied for many years. For example, change point detection analyzes when the statistical distribution of a time series changes~\cite{picard_1985}. Another related field is adaptive control, where the system parameters are continuously adjusted to account for discrepancies of expected and observed data~\cite{Isermann1991}. We focus our review on recent methods that address distribution shifts when forecasting with neural networks. \citet{kim2021reversible} employs a trainable instance-wise normalization-and-denormalization layer to address non-stationary mean and variance in time series forecasting. \citet{arik2022self} proposes Self-Adaptive Forecasting to adapt the forecasting model using test-time training by "backcasting" and using the backcast errors to signal a potential distribution shift to adjust the weights of the model before inference. AdaRNN~\cite{du2021adarnn} splits the time series history into dissimilar segments and learns importance weights to combine the RNN hidden states over these segments. Another notion of distribution shift, adversarial attacks, has been recently considered to build more robust forecasting~\citep{yoon2022robust, liu2022towards}.

Another class of approaches to time series forecasting use specific weighting schemes to weight the time series history to make predictions, for example exponential smoothing, the Non-Parametric Time Series (NPTS) Forecaster \cite{gardner1985exponential, alexandrov2020gluonts}, or stratified sampling~\cite{lu2021variance}. In contrast to these models, our approach adaptively learns the weighting scheme over the history.  Thus, our work can be seen as a specific instantiaten of weighted re-sampling (or importance sampling) for time-series forecasting. Importance sampling weights the log-likelihood in supervised learning tasks by importance weights $w(x) = p_{test}(x)/p_{train}(x)$ where $p_{test}$ and $p_{train}$ are the respective densities from which the train/test samples are drawn~\cite{NIPS2007_be83ab3e, NIPS2006_a2186aa7, wiles2022a, pmlr-v108-li20b}. Here, we propose to avoid estimating $p_{train}$ and $p_{test}$ and instead tune the parameters of a parametric re-sampling function using Bayesian optimization.

\subsection{Scenarios of distribution shift occurrence}\label{sec: shift_types}

Assumption~\ref{eq: shift_assumption} allows us to provide a precise breakdown of the specific cases of distribution shift that our model can handle. This is important because it determines the exact applicability of the work described in this paper. As Figure~\ref{fig:shift_cases} shows, we identify three domains:

\textbf{New or reoccurring shift:} When a distribution shift is new or reoccurring in the past and there is no shift between $t \in [T-c, T]$ and $t \in [T+1, T + \tau]$. In this case the adaptive sampling mechanism is expected to focus only on the shifted regions that are relevant for the current regime.

\textbf{Reverted shift:} When a shift occurred in the past but the distribution eventually reverted back to the “standard” regime. This case is complementary to the previous one: here the adaptive sampling mechanism provides a way to exclude the shifted regions. One can see the shifts in this case as regions with noisy or corrupted data: excluding them during training would guarantee robustness against noise.

\textbf{Future shift:}  When the shift occurs in the forecasting window but not in the most recent region before the forecast start date. Our adaptive sampling mechanism cannot work if we have no information regarding the distribution shift happening in the test set (no free lunch).\footnote{Note that in our setting we consider the forecasting window to be a mere continuation of the training set, which implies that the context window used at test time was part of training inputs. In this sense, our procedure cannot handle this case if and only if the shift happens solely in the forecasting range.}

\subsection{Datasets}~\label{sec: datasets} 
\begin{table*}[t]

\begin{small}
\begin{center}

\begin{tabular}{lcc|rrr}
    \toprule
    {} & Freq. & Horizon & Number of Series & Avg. Length & Number of  observations \\
    \midrule
    \covid                 &              D &               30 &                227 &                  182 &                   41,314 \\
    \electricity                  &              H &               24 &                321 &               21,044 &                6,755,124 \\
  
    \traffic        &              H &               48 &                862 &               17,496 &               15,081,552 \\
    \solar                         &              H &               24 &                137 &                7,009 &                  960,233 \\
    \taxi                          &         30 MIN &               24 &              1,214 &                1,488 &                1,806,432 \\
    \bottomrule
\end{tabular}
 \caption{Dataset statistics for the used datasets, taken from~\citet{tsbench}.}\label{tbl:datasets}
\end{center}
\end{small}
\end{table*}

We use datasets from the Monash Time Series Forecasting Repository~\cite{godahewa2021monash} and GluonTS~\cite{alexandrov2020gluonts} for our experiments. We use the same train/test splits as described in TSBench~\cite{tsbench}. See Table~\ref{tbl:datasets} for dataset statistics. The frequencies of our used datasets are 30 minutes (30 MIN), hourly (H), and daily (D).  

In particular, we use the following datasets: 

\covid{}: Daily COVID-19 deaths of different countries between January 22 and August 21 2020~\cite{dong_interactive_2020}. 

\electricity{}: Hourly household consumption data between January 2012 and June 2014~\cite{Dua:2019}.

\traffic{}: Hourly occupancy rates of freeways in the San Francisco Bay area between 2015 and 2017~\cite{Dua:2019}. 

\solar{}: Hourly power output of 137 photovoltaic power stations in the US~\cite{lstnet}.

\taxi{}: Number of taxi rides in different locations in New York sampled at 30 minute windows~\cite{taxi:2015}.

\subsection{Training Details}~\label{sec: training_details}
The parameters of the distribution over history are optimized via Bayesian Optimization \cite{frazier2018tutorial} using the recent scalable implementation provided in \texttt{Syne-Tune} \cite{salinas2022syne}. We leveraged an AWS EC2 instance with $\approx42$ cores and set the number of random initializations to 44 to allow for sufficient exploration of the parameter space. That is, the optimizer starts with 44 trials, each with a different parameter configuration, and sequentially stops some of them or instantiate new ones given the information from the Gaussian Process posterior. The optimizer stopping criterion was chosen to be the maximum number of trials taken to completion, which was set to 200. We used the CRPS used to decide which trials were stopped or continued was the mean quantile loss over the validation set.\\
\\
The lightweight architecture is supposed to be simple enough to allow frequent re-training through adaptive sampling, so that the forecasting model can quickly adapt to new distributions. Clearly, different architectures will work better with different datasets. In all our experiments with \deepar{}~\cite{salinas2020deepar}, we found that using the default two-layer LSTM with 40 hidden units for the pre-trained model, and freezing the first layer only for adaptive training yielded the best results. To clarify, this means that $\mathbf{w}_{\text{ada}} \subset \mathbf{w}_{\text{pre}}$. Both pre-training and adaptive training were done over 100 epochs with a small dropout probability equal to $0.1$. 

For \tft{}, we selected the hyperparameter setting from the TSBench~\cite{tsbench} that resulted in lowest CRPS loss on the validation set for each dataset. The hyperparameters tuned in TSBench are the context length, the number of hidden states, and the number of heads. 

For the \deepar{} and \tft{} models, we use the implementation available in GluonTS~\cite{alexandrov2020gluonts}.

\subsection{Additional Figures}
\clearpage
 \begin{figure}[t!]
     \floatbox[{\capbeside\thisfloatsetup{capbesideposition={right,top}}}]{figure}[\FBwidth]
        {\caption{\textbf{Adaptive sampling uses $p_\bphi$ to sample training windows to come from distributions that are similar to the forecasting window.} Both panels show a time series (\textcolor{Blue}{blue}) at the top, with windows (\textcolor{YellowOrange}{orange}) sampled during training, and the distribution over time steps at the bottom. \textbf{a)} Uniform sampling. \textbf{b)} Adaptive sampling.
     }
     \label{fig: ada_sampling}}
     {\includegraphics[width=0.7\textwidth]{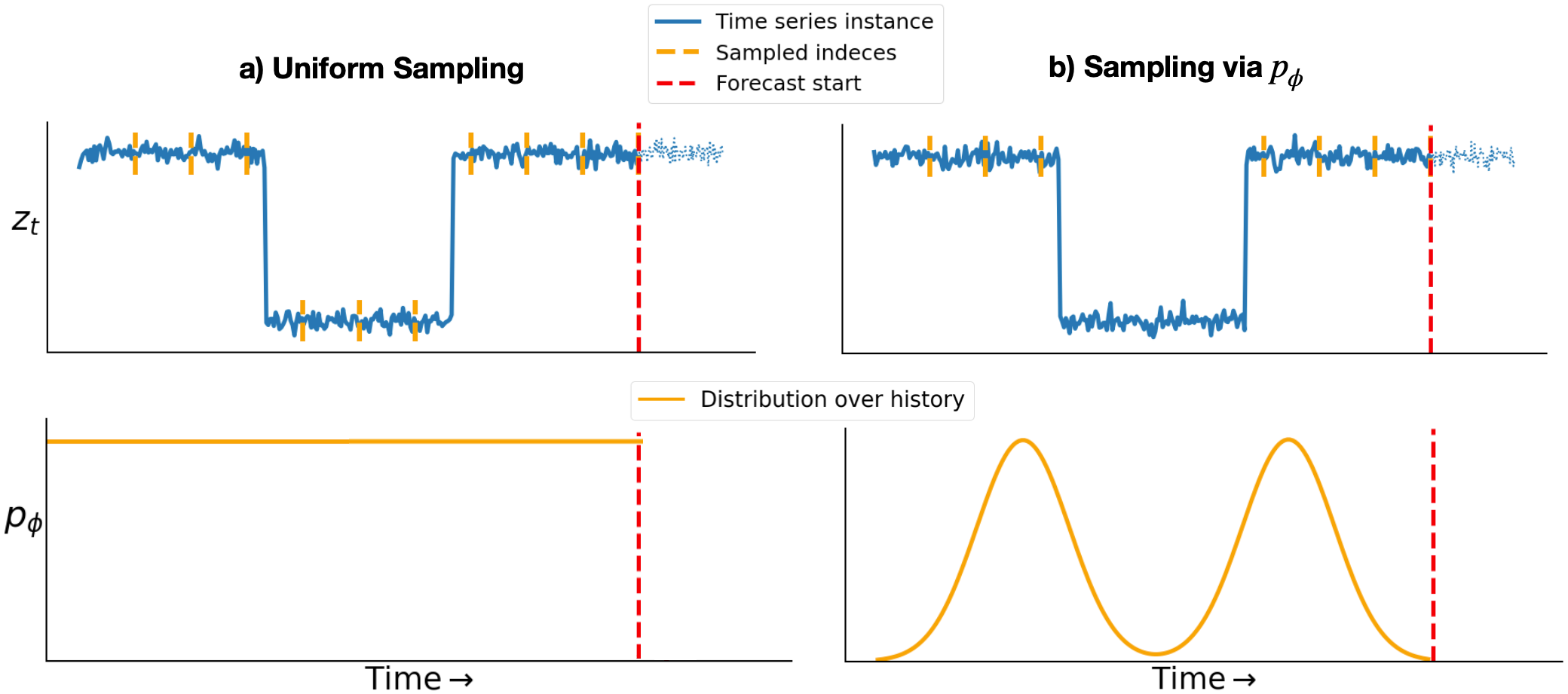}}
 \end{figure}

\end{document}